\documentclass[a4paper,twoside]{article}

\usepackage{epsfig}
\usepackage{subcaption}
\usepackage{calc}
\usepackage{amssymb}
\usepackage{amstext}
\usepackage{booktabs}
\usepackage{amsmath}
\usepackage{amsthm}
\usepackage{multicol}
\usepackage{pslatex}
\usepackage{apalike}
\usepackage[bottom]{footmisc}
\usepackage[utf8]{inputenc}
\usepackage[T1]{fontenc}
\usepackage{xspace}
\usepackage{url}
\usepackage{verbatim}
\usepackage{enumerate}
\usepackage{lmodern}
\usepackage{enumitem}
\usepackage{graphicx} 
\usepackage{longtable}
\usepackage[table,xcdraw]{xcolor}
\usepackage{multirow}
\usepackage{acronym}
\usepackage{threeparttable}
\usepackage{mathabx}
\usepackage{tabularx,ragged2e,booktabs,caption}
\usepackage{array}
\usepackage{cite}
\usepackage{textcomp}
\usepackage{academicons}
\usepackage{xcolor}
\usepackage{csquotes} 
\usepackage{algorithm2e}
\usepackage{algorithmic}
\usepackage{tikz}
\usepackage{adjustbox} 
\usepackage{hyperref}
\usepackage{SCITEPRESS}     
\usepackage{array}
\usepackage{multirow}
\usepackage{multicol}
\usepackage{booktabs}
\usepackage{listings}
\usepackage{subcaption}
\usepackage{appendix}

\usepackage[headsepline,manualmark]{scrlayer-scrpage}
\clearpairofpagestyles
\ohead{\pagemark}
\ihead{\headmark}

\newcolumntype{C}{>{\centering\arraybackslash}X} 
\newcommand{\orcid}[1]{\href{https://orcid.org/#1}{\includegraphics[width=1em]{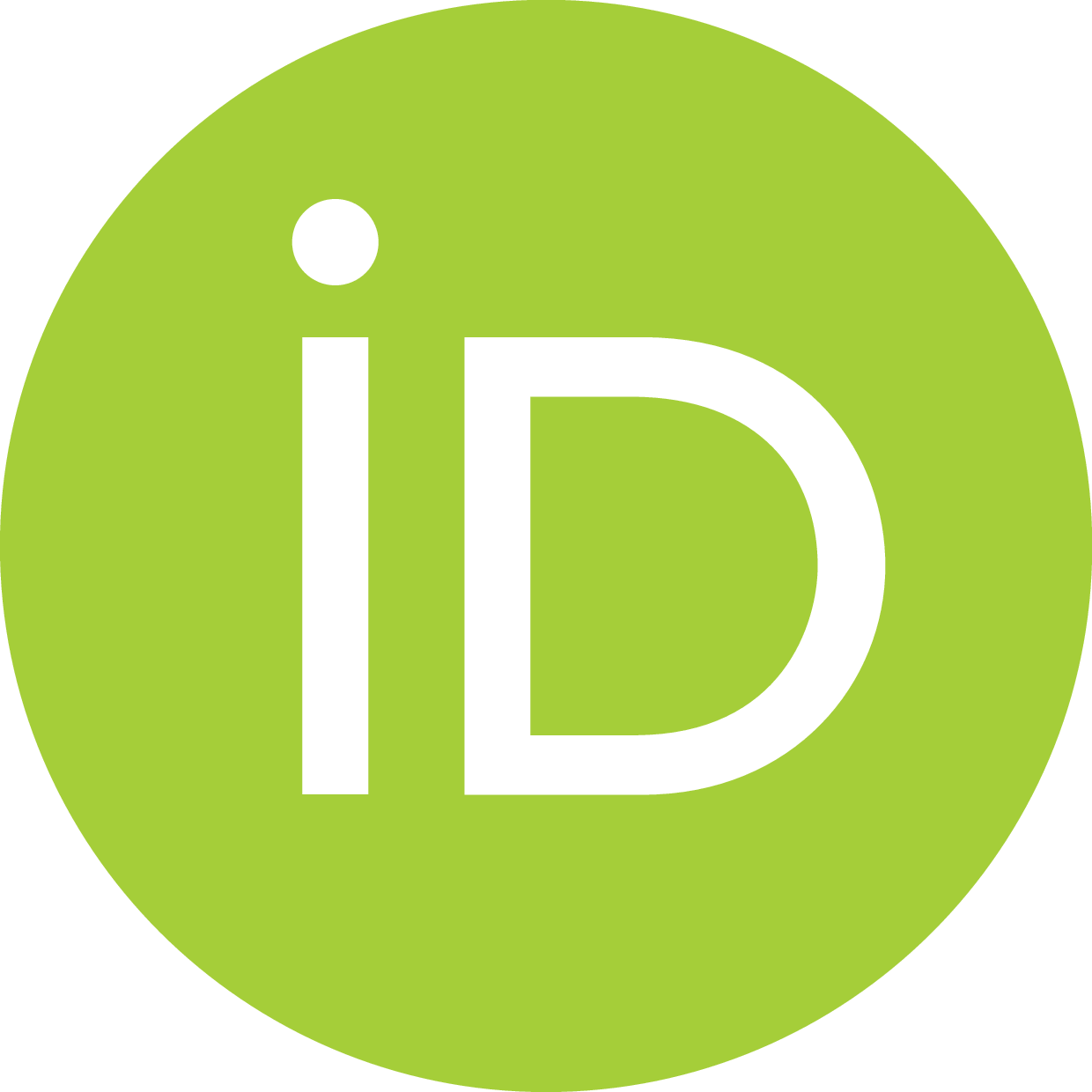}}}

\begin{document}

\title{Simulation-based Performance Evaluation of 3D Object Detection Methods with Deep Learning for a LiDAR Point Cloud Dataset in a SOTIF-related Use Case}

\author{\authorname{Milin Patel\sup{1}\orcidAuthor{0000-0002-8357-6018} and Rolf Jung\sup{2}\orcidAuthor{0000-0002-0366-4844}}
\affiliation{\sup{1}Institute for Advanced Driver Assistance Systems and Connected Mobility, Kempten University of Applied Sciences, Benningen, Germany}
\affiliation{\sup{2} Faculty of Computer Science, Kempten University of Applied Sciences, Kempten, Germany}
\email{\{milin.patel, rolf.jung\}@hs-kempten.de}
}


\abstract{Safety of the Intended Functionality (SOTIF) addresses sensor performance limitations and deep learning-based object detection insufficiencies to ensure the intended functionality of Automated Driving Systems (ADS). This paper presents a methodology examining the adaptability and performance evaluation of the 3D object detection methods on a LiDAR point cloud dataset generated by simulating a SOTIF-related Use Case. The major contributions of this paper include defining and modeling a SOTIF-related Use Case with 21 diverse weather conditions and generating a LiDAR point cloud dataset suitable for application of 3D object detection methods. The dataset consists of 547 frames, encompassing clear, cloudy, rainy weather conditions, corresponding to different times of the day, including noon, sunset, and night. Employing MMDetection3D and OpenPCDET toolkits, the performance of State-of-the-Art (SOTA) 3D object detection methods is evaluated and compared by testing the pre-trained Deep Learning (DL) models on the generated dataset using Average Precision (AP) and Recall metrics.}

\onecolumn \maketitle \normalsize \setcounter{footnote}{0} \vfill

\section{\uppercase{Introduction}}
\label{chap:introduction}
\ac{ads} utilizes object detection methods to detect and respond to objects and events in the environment. Object detection methods methods use \ac{dl} models that are trained on large datasets of images or \acs{lidar} point clouds to detect objects entities including pedestrians, vehicles, and obstacles. However, the performance of these methods is influenced by the quality of the training data, the complexity of the driving scenarios, and the inherent model constraints.  

The process of 3D point cloud object detection involves recognizing and locating objects within a three-dimensional space using point cloud data, which is a collection of data points representing the surfaces of objects in a given environment, typically obtained from \acs{lidar} sensor. \cite{Arzhanov20193DOD}  

Adapting \acs{dl} models to perform well across diverse domains and in adverse weather conditions to improve generalization is a complex task \cite{9829218}. The survey by \cite{Guo.27.12.2019} provided an overview of \acs{dl} for 3D point clouds but did not address applications in \acs{sotif}-related Use Cases. The study by \cite{8864642} evaluated \acs{lidar} object detection using simulated data but did not account for varying weather conditions essential for \acs{sotif}-related Use Cases. Although \cite{peng2022pesotif} developed a dataset focusing on visual \acs{sotif} scenarios in traffic, the application to point cloud-based object detection methods was not examined. Furthermore, \cite{abrecht2023deep} highlighted safety concerns in automated driving perception without delving into the specifics of \acs{lidar} sensor data in varying weather conditions. 

The challenges in object detection from point clouds include the assumption that object categories do not change over time, leading to performance degradation when learning new classes consecutively \cite{10054469}. Another challenge is detecting small objects due to the lack of valid points and the distortion of their structure within the point cloud \cite{huang2022psadet3d}. Moreover, the raw point cloud may be sparse or occluded, resulting in decreased detection performance \cite{10115327}. The challenges in 3D point cloud object detection include noisy points, blind and cluttered scenes, uncompleted parts of objects, and data analysis difficulties. \cite{elharrouss20233d}

The performance of 3D object detection methods with \acs{dl} can be evaluated in a simulation environment by using benchmark datasets. Simulation environments allow for the creation of synthetic datasets replicating different environmental conditions, providing a controlled and repeatable setting suitable for evaluating the performance of 3D object detection methods \cite{10140999}. Benchmark datasets provides standardized evaluation metrics, facilitating comparisons between different methods, and enabling the reproducibility of results. The KITTI dataset is a commonly used benchmark dataset for 3D object detection, offering a diverse range of object categories and annotated ground truth data. \cite{10140999},\cite{10.1117/12.2675099}

CARLA is utilized for generating LiDAR point cloud datasets for 3D object detection due to its scalable simulation environment, the ability to simulate a variety of driving scenarios, and sensor simulation. However, it also faces challenges of the simulation-to-reality gap, limited environmental diversity, and computational demands. \cite{10161226}, \cite{9999668}, \cite{r20233d}, \cite{9970914}

To prepare a point cloud dataset from SOTIF-related Use Case simulations for 3D object detection, it is crucial to define specific driving scenarios encompassing safety considerations, potential failure modes, and challenging environmental conditions. Utilizing simulation environments and synthetic data generation tools is essential for capturing point cloud data representing these scenarios, while incorporating variations in lighting, weather conditions, object occlusions, sensor noise, and other factors relevant to SOTIF-related use considerations. Additionally, annotating the simulated point cloud data with ground truth labels for objects and applying \acs{dl}-based object detection methods are critical to ensure the relevance of the dataset to the intended functionality.

Acknowledging the conceptual background established by existing literature, this paper aims to expand the application of 3D object detection methods in \acs{sotif}-related Use Case by focusing on \acs{lidar} point clouds.

\subsection{Contribution}
\label{subchap:contribution}
The main contributions of this paper are summarized as follows:
\begin{enumerate}
	
\item[(i)] Defining and modeling of a \acs{sotif}-related Use Case incorporating 21 diverse weather conditions in the CARLA simulation environment.

\item[(ii)] Simulation and generation of a \acs{lidar} point cloud dataset, suitable for the application of \ac{sota} 3D object detection methods.

\item[(iii)] Application and evaluation of point cloud based 3D object detection methods on the generated dataset using MMDetection3D and OpenPCDet toolkits, using \ac{ap} and Recall metrics.
\end{enumerate}

The work is limited to 3D object detection using \acs{lidar} point clouds. No RGB images or other sensory data are used. This paper adopts an approach where detection methods, pretrained on a real-world dataset, are evaluated against a simulated LiDAR point cloud dataset. This methodology enables a direct assessment of the \acs{dl} models' adaptability to a range of simulated conditions.

\subsection{Research Questions}
\label{subchap:research questions}
This paper aims to resolve the following research questions:
\begin{itemize}
	\item [RQ1.] What approach should be taken to prepare a dataset from a \acs{sotif}-related Use Case simulation that is suitable for point cloud-based 3D object detection methods? \label{RQ1}
	
	\item [RQ2.] Which \acs{sota} 3D object detection methods are compatible with dataset generated from a \acs{sotif}-related Use Case simulation?
	
	\item [RQ3.] How do point cloud-based 3D object detection methods perform when applied to a dataset generated from a \acs{sotif}-related Use Case simulation?
\end{itemize}

\subsection{Structure of the Paper}
Following the introduction, this paper is organized into four main chapters. Chapter \ref{chap:methodology} presents an overview of the SOTIF-related Use Case, dataset structure, the CARLA simulation environment, point cloud-based 3D object detection methods, and the evaluation metrics used. 

Chapter \ref{chap:implementation} delineates the process of dataset generation from CARLA and applying 3D object detection methods to this dataset. Chapter \ref{chap:results_discussion} presents the comparison of performance evaluation results obtained from the application of 3D object detection methods.

The paper concludes with Chapter \ref{chap:conclusion}, summarizing the work and suggesting directions for future research directions. Additionally, Appendix \ref{chap_abbreviations} lists the abbreviations used throughout the paper.

\section{\uppercase{Methodology}}
\label{chap:methodology} 
\begin{figure}[h!]
	\centering
	\includegraphics[width=\linewidth,keepaspectratio]{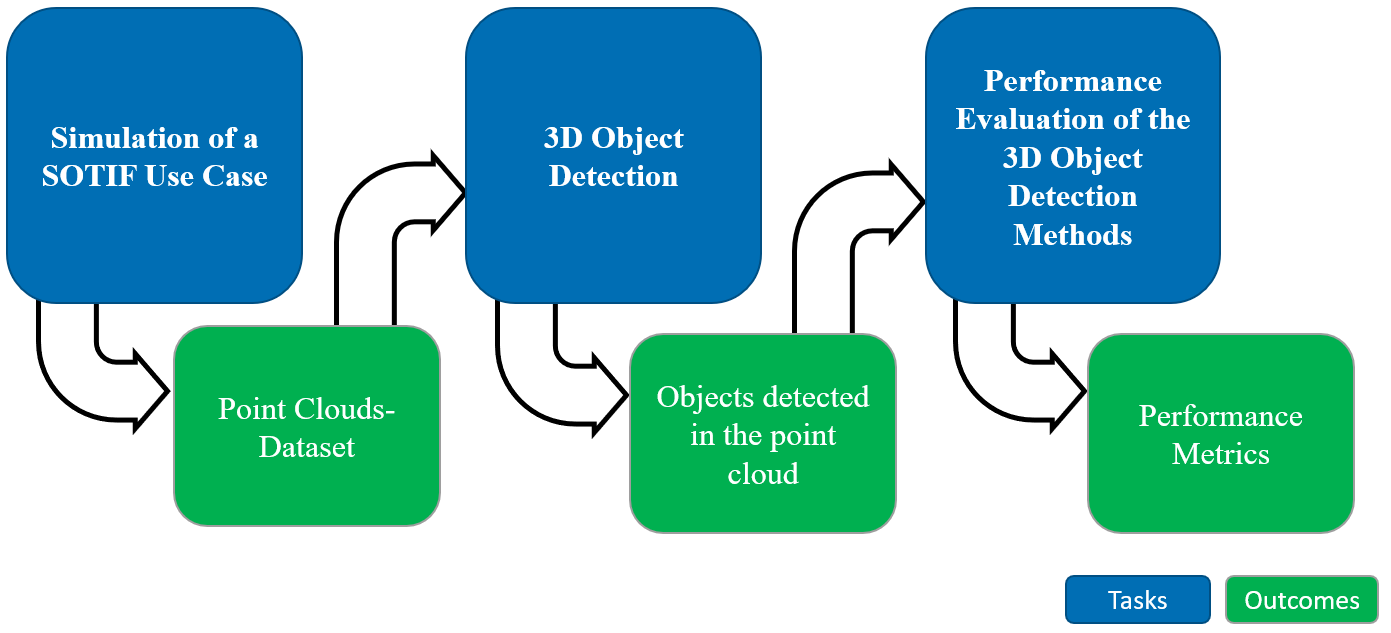}
	\caption{Schematic Representation of the Methodological Approach}
	\label{fig:methodology_overview}
\end{figure}

This chapter outlines the methodology used to evaluate the performance of 3D object detection methods when applied to a \acs{lidar} point cloud dataset from a \acs{sotif}-related Use Case. Figure \ref{fig:methodology_overview} presents a schematic representation of the methodological approach, detailing the tasks and anticipated outcomes at each stage.

The primary task involves simulating a \acs{sotif}-related Use Case incorporating diverse weather conditions to generate a point cloud dataset using the CARLA simulation environment \cite{dosovitskiy2017carla}. The generated dataset is then processed with 3D object detection methods with \acs{dl} models, which are provided and supported by the MMDetection3D and OpenPCDet toolkits \cite{MMDetection3D.2020, openpcdet2020}, customized for \acs{lidar}-based 3D object detection. These toolkits facilitate the performance evaluation of 3D object detection methods on the generated dataset against established performance metrics. 

\subsection{Description of the \acs{sotif}-related Use Case}
\label{subchap:sotif_description}
\begin{figure}[!h]
	\centering \includegraphics[width=\linewidth,keepaspectratio]{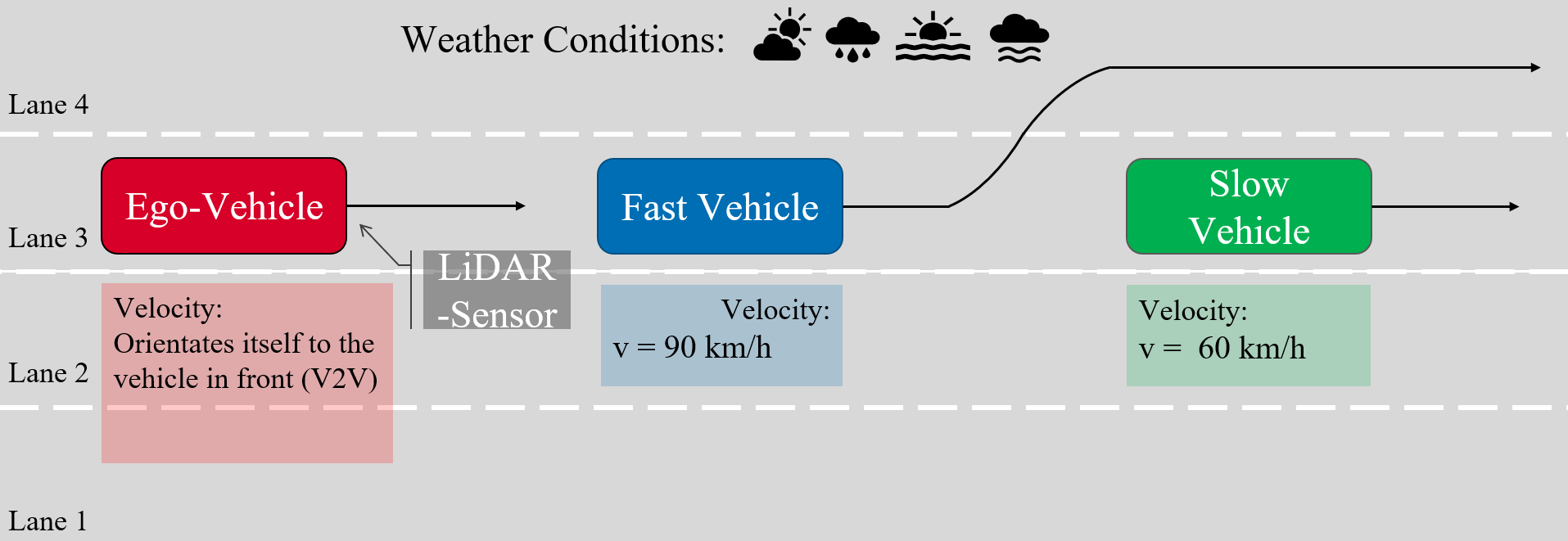}
	\caption{Description of the \acs{sotif}-related Use Case}
	\label{fig:Use case}
\end{figure}

In the \acs{sotif}-related Use Case depicted in Figure \ref{fig:Use case}, the focus is on the \enquote*{Ego-Vehicle} equipped with a \acs{lidar} sensor that operates on a multi-lane highway. The \acs{lidar} sensor's functionality is to generate detailed three-dimensional point clouds of the Ego-Vehicle's surrounding. This enables the Ego-Vehicle to detect surrounding vehicle, measure relative distance and velocity, and adjust its own velocity under diverse weather conditions.
The Ego-Vehicle is positioned in lane 3 and is programmed to adapt its velocity based on the traffic flow, specifically the vehicles directly ahead. It utilizes \ac{v2v} communication to estimate the velocity and distance to the preceding vehicle, termed as the \enquote*{fast vehicle}, traveling at 90 km/h. This enables the Ego-Vehicle to adjust its velocity accordingly, maintaining a safe following distance without requiring driver intervention.

Subsequently, the fast vehicle changes lanes to overtake a \enquote*{slow vehicle}, traveling at 60 km/h in the same lane as the Ego-Vehicle. This maneuver necessitates the Ego-Vehicle's \acs{lidar} sensor to promptly detect the slower vehicle and execute a deceleration maneuver to prevent a potential collision.

Furthermore, the Use Case includes diverse weather conditions as outlined in the Table \ref{table:weather_presets}, range from clear conditions optimal for \acs{lidar} sensor performance to challenging weather conditions like rain and cloud cover that could impede visibility and \acs{lidar} sensor performance. 

Table \ref{table:weather_presets} categorizes weather conditions into three columns corresponding to different times of day: Noon, Night, and Sunset.

\begin{table}[h!]
	\centering
	\caption{Diverse Weather Conditions for \acs{sotif}-related Use Case Dataset Generation, \cite{.17.01.2024}}
	\label{table:weather_presets}
	\begin{adjustbox}{max width=7cm}
		\begin{tabular}{@{}|l|l|l@{}|}
				 \hline
		\textbf{Noon}       & \textbf{Night}         & \textbf{Sunset}          \\ \hline
		ClearNoon           & ClearNight             & ClearSunset              \\ \hline
		CloudyNoon          & CloudyNight            & CloudySunset             \\ \hline
		WetNoon             & WetNight               & WetSunset                \\ \hline
		WetCloudyNoon       & WetCloudyNight         & WetCloudySunset          \\ \hline
		MidRainyNoon        & MidRainyNight          & MidRainSunset            \\ \hline
		HardRainNoon        & HardRainNight          & HardRainSunset           \\ \hline
		SoftRainNoon        & SoftRainNight          & SoftRainSunset           \\ \hline
		\end{tabular}
	\end{adjustbox}
\end{table}

Each row of the table represents a set of weather conditions simulated during the respective time period. The \enquote*{Clear} conditions serve as a control or baseline, \enquote*{Cloudy} conditions introduce diffused lighting challenges, \enquote*{Wet} conditions incorporate reflective lane surfaces, and \enquote*{Rainy} conditions mimic visibility reduction due to precipitation. \enquote*{WetCloudy} conditions combine moisture and diffused light challenges, representing a more complex environment for \acs{lidar} sensor.

\subsection{Dataset structure}
\label{subchap:dataset_structure}
For the effective application and evaluation of 3D object detection methods on generated \acs{lidar} point cloud data dataset, the data must be accurately structured and annotated. The dataset requires label data, including precise object positions, dimensions, and classifications, collectively known as \enquote*{ground truth} data.

The dataset format and structure must align with benchmark standards to ensure compatibility with 3D object detection methods. The KITTI dataset \cite{Geiger.2012}, established by the Karlsruhe Institute of Technology and the Toyota Technological Institute in 2012, has become the benchmark in the autonomous driving domain. The KITTI dataset is one of the most frequently cited datasets in the research field of autonomous driving~\cite{Yin.2017} and is used by the majority of 3D object detection methods \cite{Geiger.2012}.

The generated dataset for this paper, therefore, adopts the KITTI format to facilitate the application of \acs{sota} 3D object detection methods.

\begin{figure}[h!]
	\centering \includegraphics[width=\linewidth, keepaspectratio]{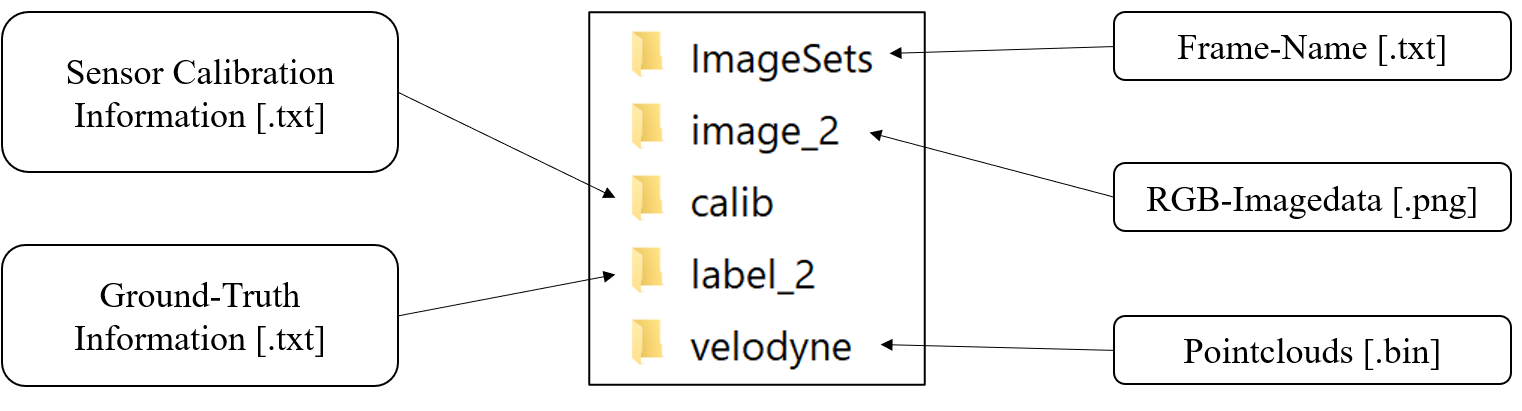}
	\caption{Generated dataset structure} 
	\label{fig:dataset_structure_folder}
\end{figure}

In Figure \ref{fig:dataset_structure_folder} illustrates the structure of the generated dataset in KITTI format. 

The subsequent description details the contents of each folder, describing their specific roles in the dataset:
\begin{description}
\item[Point Cloud Data:] The \enquote*{velodyne} folder contains point cloud data, represented as binary files (.bin).
	
\item[Image Data:] RGB images, stored as portable network graphics (.png) in the \enquote*{image\_2} folder, supplement the point cloud data by providing visual context.

\item[Calibration Data:] Stored in the \enquote*{calib} folder as text files (.txt), calibration data provide parameters for camera sensor alignment and calibration. 

\item[Frame Names:] Recorded frame identifiers (e.g., 000001, 000002, 000003) are documented in a text file (.txt) within this folder. These identifiers facilitate dataset division into training, validation, and testing subsets.

\item[Label Data:] The \enquote*{label\_2} folder contains ground truth data in a text file (.txt), encompassing  details of object class, dimensions, and bounding box coordinates. This data is pivotal for validating object detection accuracy and training \acs{dl} models.
\end{description}

\subsection{Simulation Environment : CARLA}
\label{subchap:simulation_carla}
CARLA, an open-source platform, was selected for its capability to simulate intricate traffic scenarios with customization weather conditions and provides diverse suite of perception sensors, including Camera, Radar and \acs{lidar}. The versatility make CARLA an ideal choice for generating a custom dataset for \acs{lidar}-based 3D point cloud detection application within controlled simulation environment. \cite{dosovitskiy2017carla}

\begin{figure}[h!]
	\centering
\includegraphics[width=\linewidth,keepaspectratio]{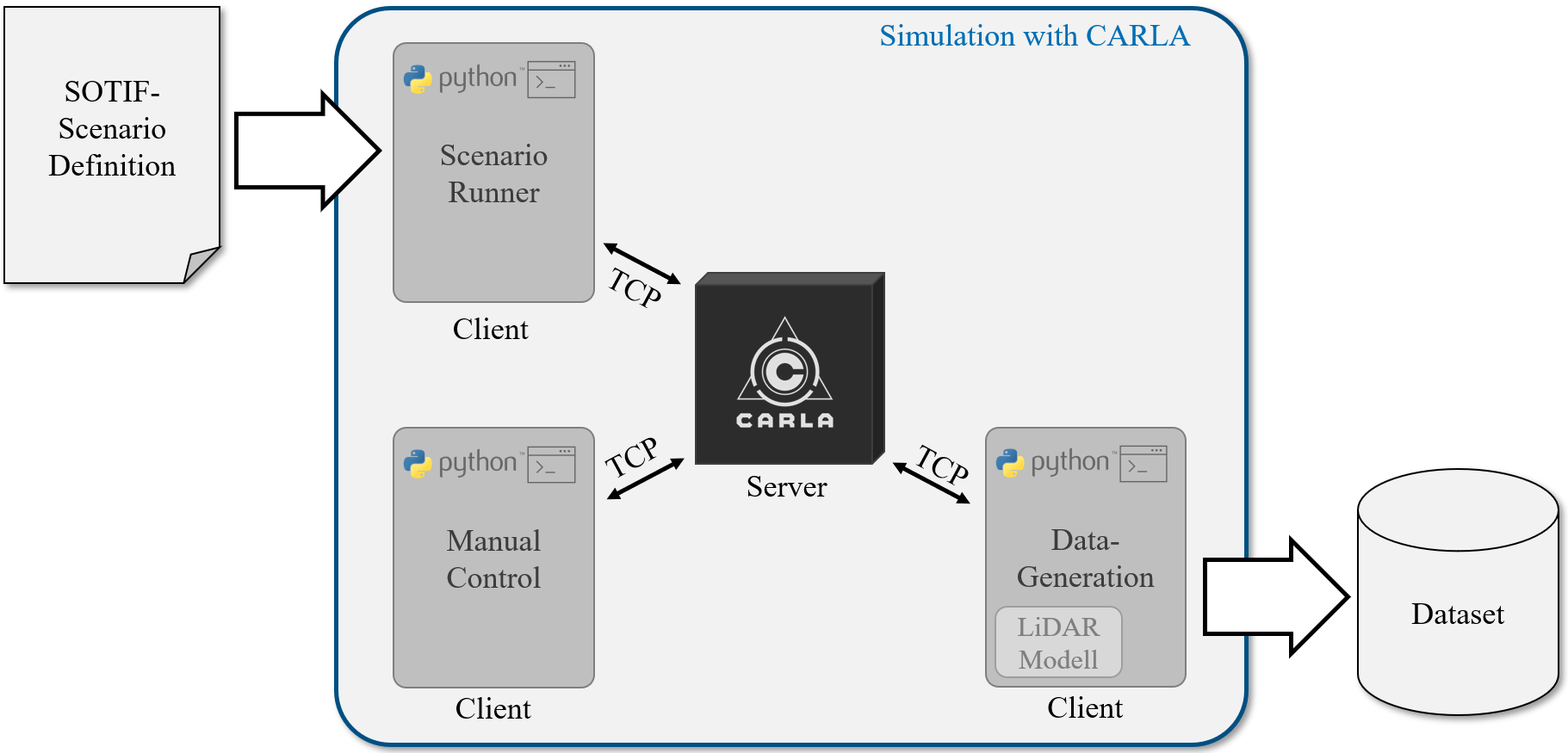}
	\caption{CARLA Simulation Environment Framework}
	\label{fig:simulation}
\end{figure}

Figure \ref{fig:simulation} depicts the server-client framework foundational to CARLA's functionality.

The CARLA server is the core of the virtual environment where simulations are hosted. The simulation can be customized and controlled as required via \mbox{clients}. A client is started by executing a Python script. Communication between clients and the server is established over a \ac{tcp}. 

As shown in Figure \ref{fig:simulation}, three \mbox{clients} are used for the simulation. The \enquote*{Scenario Runner} client sets the stage for the \acs {sotif}-related Use Case by utilizing a configuration file that specifies the weather conditions, virtual map layout, and the vehicle dynamics.

Simultaneously, the \enquote*{Manual Control} client, linked to the \enquote*{Scenario Runner} for initializing the Ego-Vehicle's characteristics, allows for independent vehicle control outside the Scenario Runner's domain.

The \enquote*{Data-Generation} client systematically structures the data storage folders, converts the simulation output into the required formats, and compiles the data. The resulting dataset, as described in Chapter \ref{subchap:dataset_structure}, is prepared for subsequent data processing in 3D object detection application.

\subsection{Point cloud-based 3D object detection methods}
\label{subchap:pointcloud3dobjectdetection}
\begin{figure}[h!]
	\centering 
	\includegraphics[width=\linewidth,keepaspectratio]{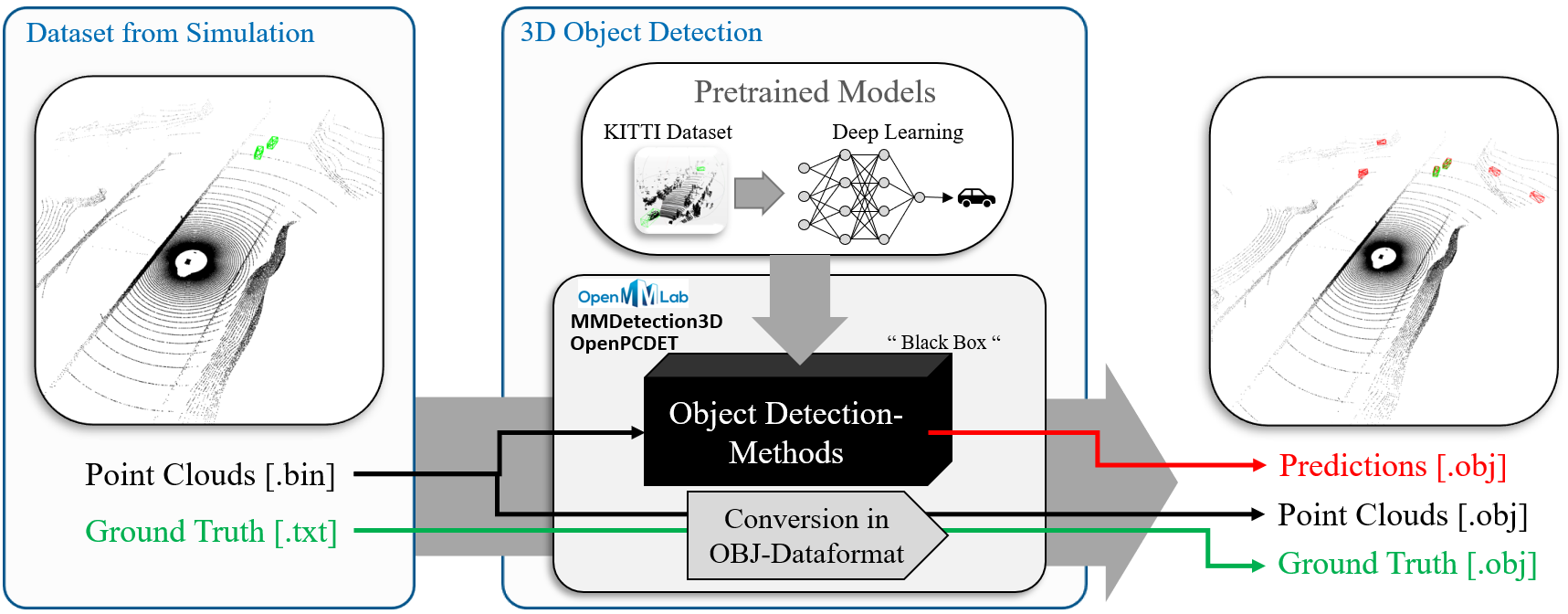}
	\caption{Overview of point cloud-based object detection}
	\label{fig:object detection}
\end{figure}

The workflow for 3D object detection is presented in the accompanying Figure \ref{fig:object detection}.

The simulation produces a dataset comprising point clouds, files in binary format encoding the 3D spatial data collected by \acs{lidar} sensors. Ground truth data, in text format, providing object location and size within the point clouds. 

3D object detection methods that utilize \acs{dl} models are employed to process the simulation-based generated dataset as input. The methodologies, operating as a \enquote{Black Box}, transform the input data into output predictions, represented by bounding boxes around detected objects within the point clouds. Two toolkits were employed: MMDetection3D and OpenPCDet. 

\begin{figure}[!h]
	\includegraphics[width=\linewidth,keepaspectratio]{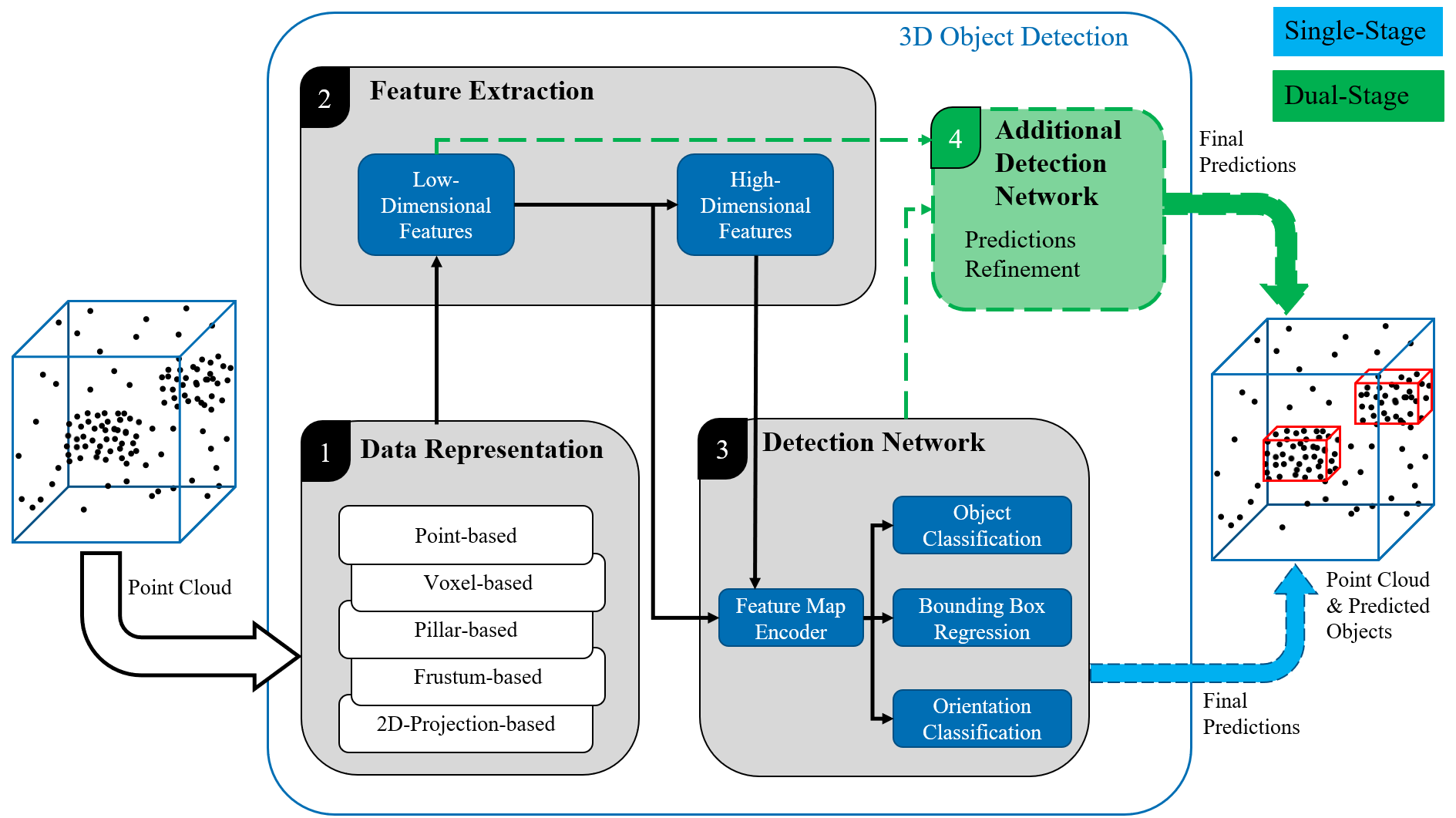}
	\caption{General structure of a 3D Object Detection model according to \cite{Fernandes.2021}}
	\label{fig:struct_objectdetection}
\end{figure}

\acs{dl}-based 3D object detection methods can be described by a uniform architecture. The structure consists of three blocks \cite{Fernandes.2021}:

\begin{itemize}
	\item Data Representation
	\item Feature Extraction
	\item Detection Network
\end{itemize}
Figure \ref{fig:struct_objectdetection} shows the general structure of point cloud-based 3D object detection methods.

\subsubsection{Data Representation }
Data representation refers to the process of transforming raw data into a structured format that can be efficiently processed by \acs{dl} models. There are several types of data representation approaches, including:  \cite{Fernandes.2021}.

\begin{description}
\item \textbf{Point-based:} The point cloud is processed directly. For each point, a feature vector is extracted, in which the neighboring properties are added. The low-dimensional properties of the individual points are combined into high-dimensional properties. \cite{Fernandes.2021}, \cite{DBLP:journals/corr/QiSMG16}
	
\item \textbf{Voxel-based:}
The point cloud is segmented into uniformly sized voxels, representing values on a 3D grid, allowing for aggregation of points within each voxel \cite{Fernandes.2021}, \cite{Zhou.2017}.
	\begin{figure}[h!]
		\centering 
		\includegraphics[width=\linewidth,keepaspectratio]{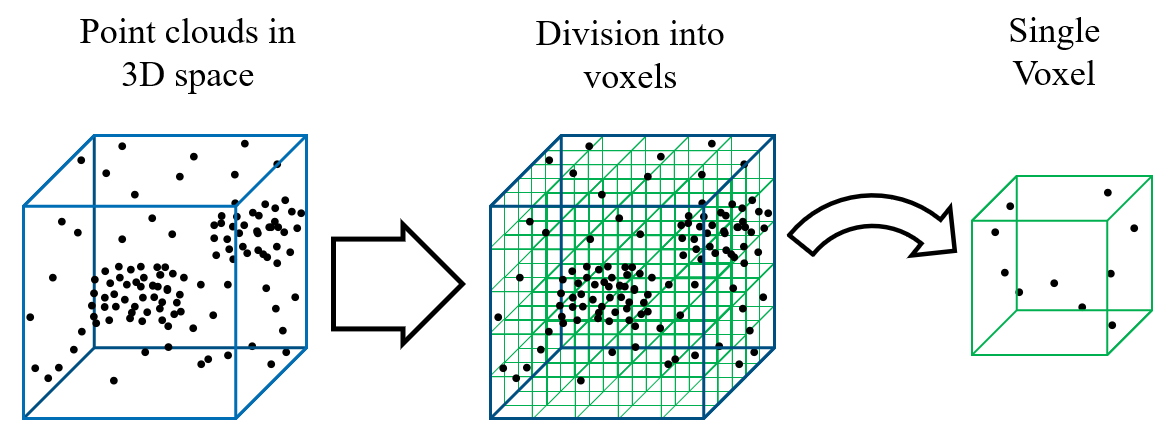}\caption{Division of the 3D space into voxels}
		\label{fig:voxelisierung}
	\end{figure}
	
\item \textbf{Pillar-based:}
Point clouds are divided into vertical columns pillars that are located on a uniform 2D grid ($x$-$y$ plane). The height of the column is equal to the number of points above the corresponding grid position. \cite{Fernandes.2021}, \cite{lang2019pointpillars}
	
	\begin{figure}[h!]
		\centering \includegraphics[width=0.7\linewidth, keepaspectratio]{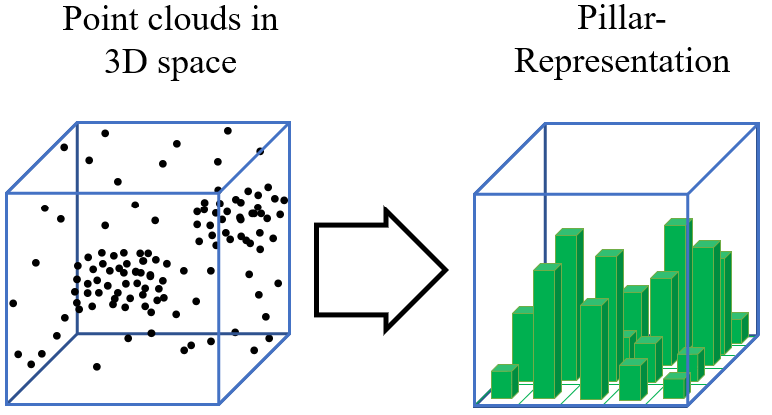}
		\caption{Transferring point cloud information into pillars}
		\label{fig:pillarbased}
	\end{figure}
	
\item \textbf{frustums-based:} Point clouds are converted into frustums. To do this, an image-based object detection method is first applied. In an image, a possible region in which an object can be located is detected. The image is then projected into the three-dimensional point cloud, creating a truncated cone.  \cite{Fernandes.2021}, \cite{Qi.2018}.
	
	\begin{figure}[h!]
		\centering 
		\includegraphics[width=\linewidth, keepaspectratio]{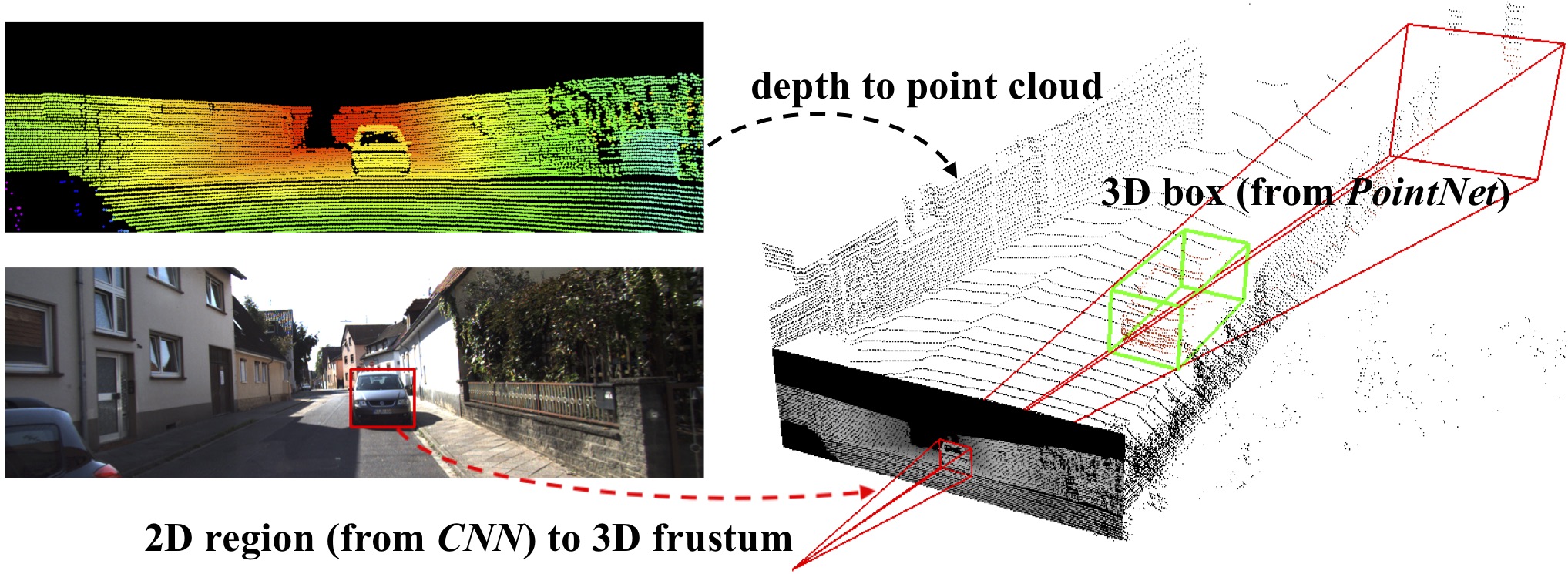}
		\caption{Generation of a frustum representation on the basis of an image \cite{Qi.2018}}
		\label{fig:frustumbased}
	\end{figure}
\item \textbf{2D-Projections-based:} Three-dimensional \acs{lidar} point cloud data is converted into a 2D projection. Different projection schemes can be used: frontal view, area view, or bird's-eye view \cite{Fernandes.2021}.
\end{description}

Point-based and voxel-based methods are computationally intensive than methods in which the data is transformed from three-dimensional to two-dimensional space. \cite{Fernandes.2021}.

\subsubsection{Feature Extraction}
It extracts low- and high-dimensional features to create a \textit{feature map}.
Optimised feature extraction is the basis for subsequent object classification at a later stage \cite{Fernandes.2021}. A distinction is made between different methods for feature extraction:

\begin{description}
	\item \textbf{Point-by-point feature extraction:}
 The whole point cloud is processed to extract features individually from points, subsequently creating a collective high-dimensional feature set \cite{Fernandes.2021}.

\item \textbf{Segment-wise feature extraction:}
Segmentation of the point cloud precedes point-by-point classification, improving efficiency and reducing computational load \cite{Fernandes.2021}.

\item \textbf{Object-by-object feature extraction:}
Initial 2D object detection is performed to create 3D bounding boxes, thus narrowing the search to regions containing objects of interest \cite{Fernandes.2021}.
 
\item \textbf{\acp{cnn}:}
Point clouds are inherently sparse, making direct 3D data processing time-consuming and computationally intensive. The inherent sparsity of point clouds is leveraged by \acp{cnn} to reduce the number of points considered \cite{Fernandes.2021}.

\end{description}

\subsubsection{Detection Network} 
It includes the determination of object classes, bounding box regression, and orientation assessment, occasionally considering object velocity \cite{Fernandes.2021}.

\begin{description}
	\item \textbf{\item Detector architectures:}
An additional detection network is used, with which the detection of the data is carried out, as shown in  Figure \ref{fig:struct_objectdetection}.
	\begin{itemize}
		\item single stage: Utilizes a singular network for detection.
		\item dual stage: Employs an additional network to refine predictions.
	\end{itemize}
\end{description}
Dual-stage detectors are typically more accurate but slower, while single-stage detectors are more efficient and less time-consuming, which makes them easier to apply to real-time systems  \cite{Fernandes.2021}.

\begin{description}
	\item \textbf{Detection module techniques:}
	\begin{itemize}
		\item \textbf{region proposal based:}
Generates \ac{roi} using a low-level algorithm. These proposed regions are then further processed and analyzed by subsequent stages of the detection system to determine the presence and characteristics of objects within those regions.
	\item \textbf{anchorless detectors:}
Suitable for occluded or truncated objects, these detectors do not pre-define candidate regions.
	\end{itemize}
\end{description}

Data conversion to OBJ format standardizes evaluation by providing a consistent format for comparing predictions with ground truth locations.

\subsection{Performance Evaluation}
\label{subchap:performance_evaluation}
The selection of appropriate performance evaluation metrics for 3D point cloud-based object detection methods depends on the characteristics of the point cloud data and the specific context. As dataset is prepared in KITTI format, \acs{ap} and recall are used as performance evaluation metric, aligning with established benchmark of MMDetection3D and OpenPCDet toolkits for applicability of the results.

3D object detection involves encapsulating objects within predicted bounding boxes, using point cloud data as input. A predicted bounding box for each object is represented by a vector of its center coordinates, dimensions, orientation, and class: \cite{Skoog.}
 \[
 B = [x, y, z, l, w, h, \theta, \texttt{class}]
\]
The coordinates $(x, y, z)$ denotes the center position of the object within the three-dimensional space. The dimensions of the object are encapsulated by $(l, w, h)$, outlining the extent of the bounding box that encloses it. The orientation of this bounding box is conveyed through $\theta$, which represents its rotation around the $Y$-axis in camera sensor coordinates. Additionally, the \texttt{class} parameter specifies the type of the object, differentiating between multiple entities within the dataset. Notably, in the dataset generated from the \acs{sotif}-related Use Case, the \texttt{class} parameter is uniformly \enquote*{Car}. \cite{Skoog.}
	
A confusion matrix presented in Table \ref{table:confusion_matrix_dl} is pivotal in evaluating the performance of \acs{dl} models for object detection. This matrix is composed of four elements, reflecting the model's detection accuracy against the ground truth. These elements are: \ac{TP}, \ac{FP}, \ac{TN}, and \ac{FN}. \cite{Sharath.2021}

\begin{description} 
	\item \textbf{True Positive (TP):} Correctly detected objects by the \acs{dl} model that are present in the ground truth.
	
	\item \textbf{False Positives (FP):} Objects detected by the \acs{dl} model that do not exist in the ground truth.
	
	\item \textbf{True Negatives (TN):} Correct non-detection, where the \acs{dl} model correctly identifies that no object is present in the ground truth.
	
	\item \textbf{False Negatives (FN):} objects that are present in the ground truth but are not detected by the \acs{dl} models.
\end{description}
\begin{table}[h!]
	\renewcommand{\arraystretch}{2.5}
	\centering
	\caption{Confusion Matrix for Object Detection using \acs{dl} Model, adapted from \cite{Sharath.2021} }
	\label{table:confusion_matrix_dl}
	\begin{adjustbox}{max width=7cm}
	\begin{tabular}{|c|c|c|}
		\hline
		\multicolumn{2}{|c|}{\textbf{\acs{dl} Model Prediction}} & \multirow{2}{*}{\textbf{Ground Truth}} \\ \cline{1-2}
		\textbf{Object Detected} & \textbf{No Object Detected} &  \\ \hline
		TP (True Positive) & FN (False Negative) & Object Present \\ \hline
		FP (False Positive) & TN (True Negative) & No Object Present \\ \hline
	\end{tabular}
\end{adjustbox}
\end{table}

Precision and Recall are two metrics derived from the confusion matrix. Precision is the ability of a \acs{dl} to identify only relevant objects. It is the percentage of correct positive predictions. While Recall expresses the ratio of correctly detected objects to the total number of existing ground-truth objects of a class.
\begin{equation}
	\text{Precision} = \frac{TP}{TP + FP}
\end{equation}

\begin{equation}
	\text{Recall} = \frac{TP}{TP + FN}
\end{equation}

There's a trade-off between Precision and Recall. Predicting many bounding boxes may yield high Recall but low Precision. Conversely, predicting only certain boxes may result in high Precision but low Recall. \cite{9145130}

\acs{ap} is a metric obtained using a Precision-Recall curve and is the equivalent to the integral of the Precision as a function of Recall. The \acs{ap} metric is usually averaged over a set of classes, and it provides a measure of the model's Precision-Recall performance across different thresholds.\cite{9145130}

\ac{iou} quantifies the overlap between predicted and ground truth bounding boxes. An \acs{iou} threshold is used to classify predictions. A prediction is considered correct if \acs{iou} is greater than or equal to threshold and incorrect otherwise. \cite{9145130}
\begin{equation} 
	IoU = \frac{\text{Area of Overlap}}{\text{Area of Union}}
\end{equation}
For the KITTI dataset, the \acs{ap} is calculated at an \acs{iou} threshold of $0.70$, signifying the required overlap between predicted and ground truth bounding boxes. \acs{ap} is further evaluated using 11-point and 40-point interpolation methods ($AP_{11}$ and $AP_{40}$).

$AP_{11}$ is calculated by interpolating Precision at 11 equally spaced Recall levels. The Recall levels for $AP_{11}$ are $\{0.0, 0.1, \ldots, 1.0\}$. At each Recall level $R$, the maximum Precision from all Recall levels greater than or equal to $R$ is considered. The average of these maximum Precision values yields $AP_{11}$. \cite{9145130}
\begin{equation}
	AP_{11} = \frac{1}{11} \sum_{R \in \{0.0, 0.1, \ldots, 0.9, 1.0\}} P_{\text{interp}}(R),
\end{equation}
Similarly, $AP_{40}$ is calculated using 40 Recall levels, providing a finer granularity in the Precision-Recall curve.

\begin{equation}
	AP_{40} = \frac{1}{40} \sum_{R \in \{0.025, 0.05, \ldots, 0.975, 1.0\}} P_{\text{interp}}(R),
\end{equation}
For both $AP_{11}$ and $AP_{40}$, $P_{\text{interp}}(R)$ represents the interpolated Precision at Recall level $R$, defined as:

\begin{equation}
	P_{\text{interp}}(R) = \max_{\tilde{R}: \tilde{R} \geq R} P(\tilde{R}),
\end{equation}
where $\tilde{R}$ ranges over all Recall levels greater than or equal to $R$.\cite{9145130}

OpenPCDet toolkit facilitates a evaluation of dual-stage detectors by using Recall metrics at \acs{iou} thresholds, specifically 0.30 and 0.50. The Recall metric is calculated after the entire dual-stage detection process and indicates how effectively the model detects and classifies objects, at specified \acs{iou} thresholds.

In dual-stage detectors, as explained in chapter \ref{subchap:pointcloud3dobjectdetection}, the first stage involves the \ac{rpn}, which identifies potential object locations. 

The Recall at this stage, referred to as \enquote*{Recall at \acs{roi}}, is calculated post-\acs{rpn} stage and signifies the the rate at which the \acs{rpn} successfully identifies actual objects within its proposed areas, adhering to the specified \acs{iou} thresholds. The subsequent stage refines the \acs{rpn}'s initial proposals, sharpening the object boundaries and accurately classifying each object, and is assessed by the \enquote*{Recall at \acs{rcnn}}. \cite{shi2019pointrcnn}, \cite{openpcdet2020}

\section{Implementation}
\label{chap:implementation}

This chapter details the implementation of the methodology defined in the chapter \ref{chap:methodology}, specifically focusing on the generation of a \acs{lidar} point cloud dataset using the CARLA simulation environment, the application of \acs{sota} 3D object detection methods, and the subsequent performance evaluation.

\subsection{Dataset Generation from CARLA}
\label{subchap:dataset_generation}
Utilizing the CARLA Scenario Runner extension, the \acs{sotif}-related Use Case is modeled following the procedure outlined in Chapter \ref{subchap:simulation_carla}. \cite{.12.09.2023}

Using the CARLA Scenario Runner extension, the \acs{sotif}-related Use Case was modeled, adhering to the procedure specified in Chapter \ref{subchap:simulation_carla}. This setup facilitated the definition of environmental parameters, vehicle behavior, and the arrangement of objects within the simulation.

\begin{figure}[h!]
	\centering \includegraphics[width=\linewidth,keepaspectratio]{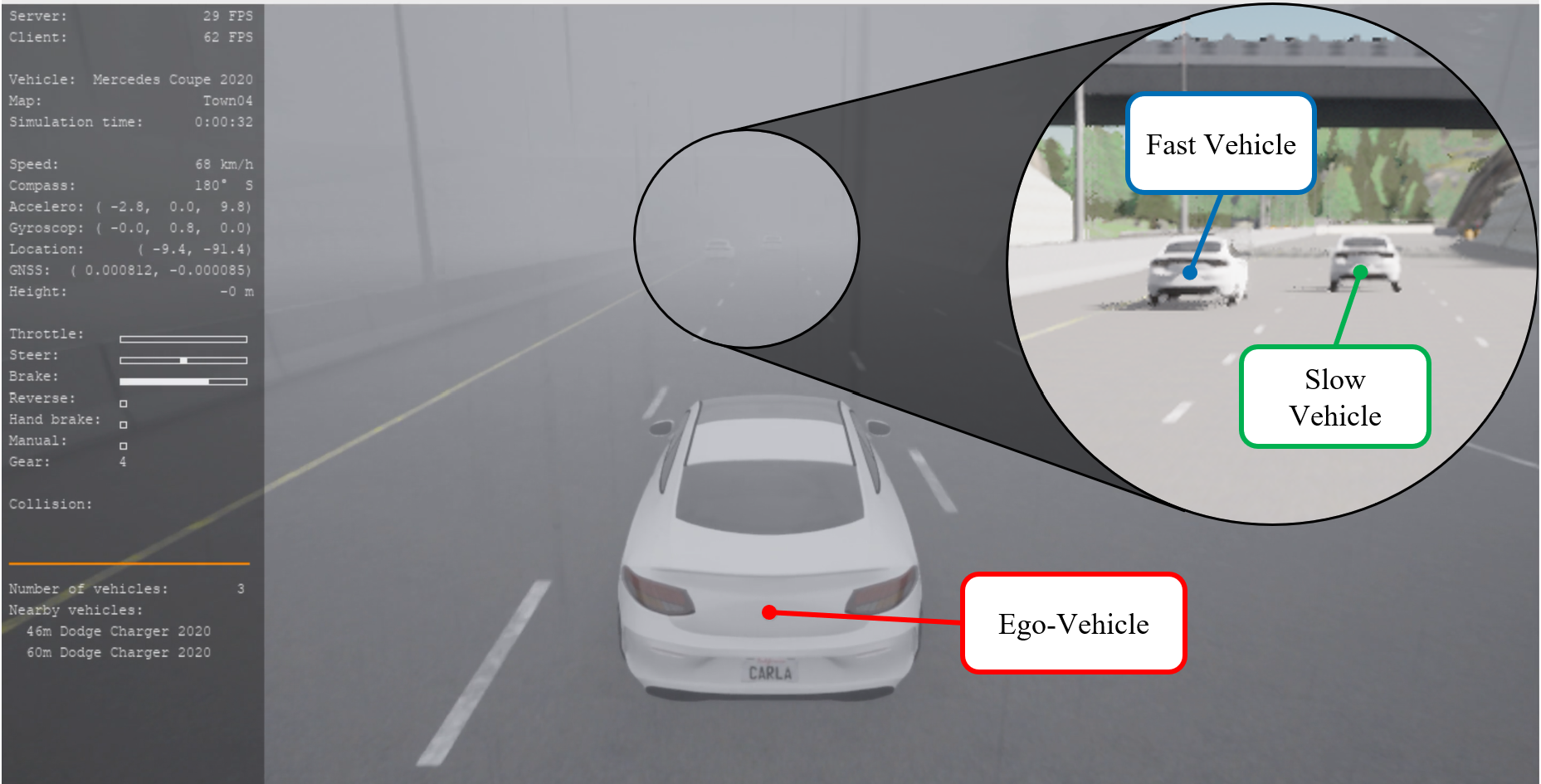}
	\caption{Simulation of the SOTIF-related Use Case in CARLA}
	\label{Fig:simulation_carla}
\end{figure}

The simulation parameters were set through an XML configuration file, detailing the virtual map, vehicle models, and weather settings. Medium-sized vehicle models with light colors were chosen for \acs{lidar} sensor visibility, with the Dodge Charger selected for both the lead and following vehicles, while the Mercedes Coupe was used for the Ego-vehicle, as shown in Figure \ref{Fig:simulation_carla}.

Python scripts, using the pyTree package, were employed to define the driving scenario's behavior, including vehicle maneuvers and positioning. A separate CARLA client was tasked with data generation, structured akin to the KITTI dataset. The script for dataset generation from \cite{Bai.2022} served as the reference, which was modified to adapt the SOTIF Use Case description.

Each data frame included point cloud data and synchronized ground-truth and image data, with frame numbering following a sequential pattern, recording every fifth frame to capture significant differences in the 10Hz \acs{lidar} data. Maintaining synchronization between server and data collection client was essential. In synchronous mode, the server advances time-steps only after all clients have completed their tasks.

A significant consideration was the difference in coordinate systems between CARLA and KITTI datasets. CARLA uses a left-handed coordinate system, while KITTI uses a right-handed system, necessitating adjustments when saving ground truth bounding box coordinates.
\begin{figure}[h!]
	\centering \includegraphics[width=\linewidth, keepaspectratio]{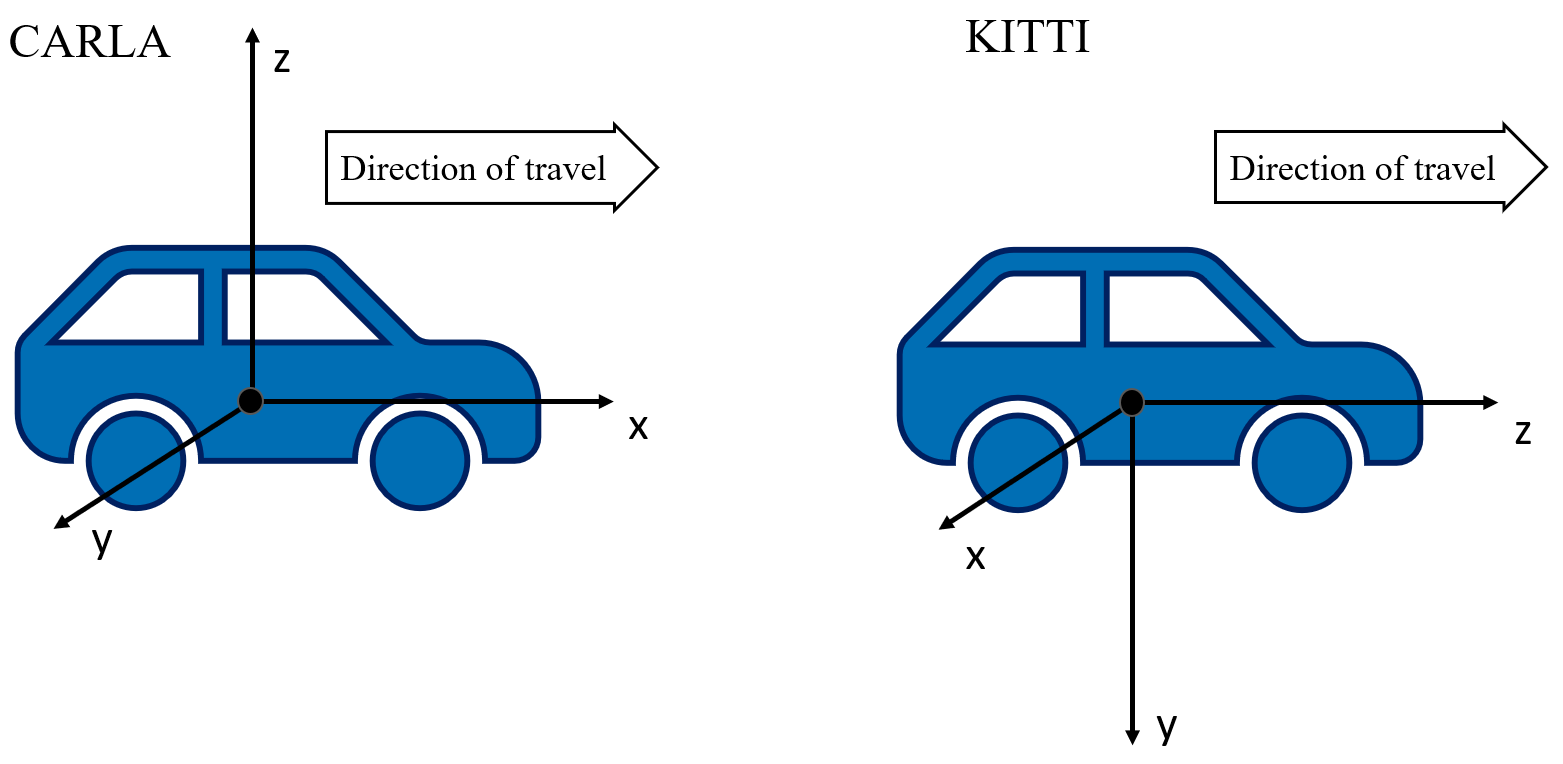}
	\caption{Comparative illustration of CARLA and KITTI coordinate systems}
	\label{fig:coordinate_systems}
\end{figure}

Before applying 3D object detection methods, the dataset was preprocessed to reorganize ground-truth data and create necessary files in Pickle (.pkl) format for bounding box processing. Employing pre-trained models from MMDetection3D and OpenPCDet, object detection was executed, enabling the comparison of predicted bounding boxes with the ground truth.

\begin{figure}[h!]
	\centering \includegraphics[width=7cm, height= 12cm]{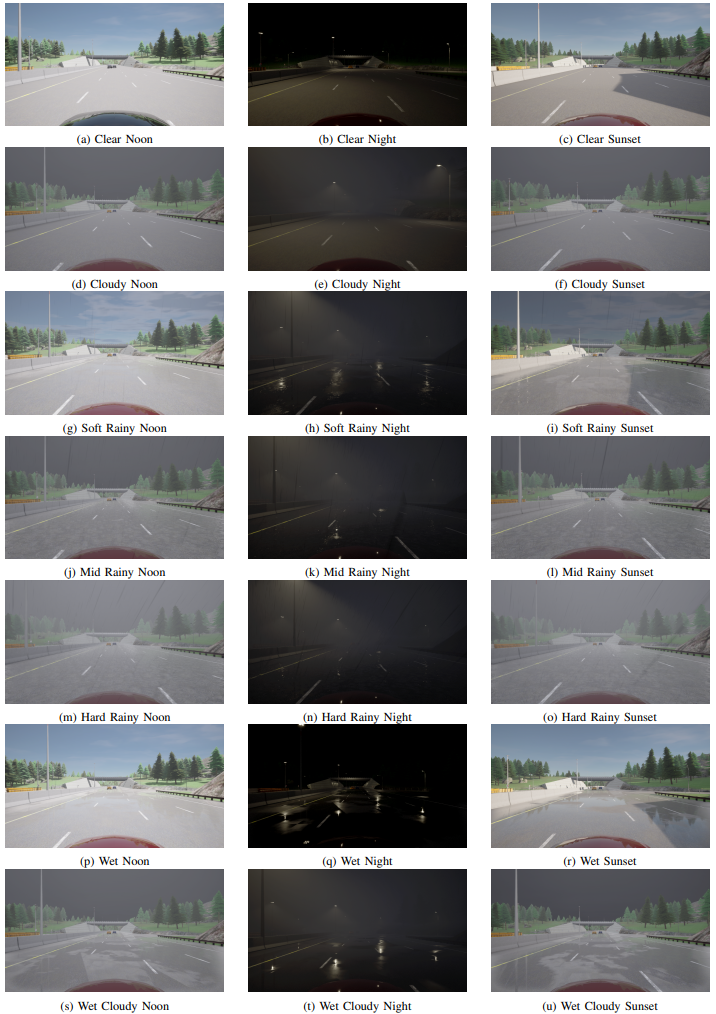}
	\caption{Overview of the generated dataset from \acs{sotif}-related Use Case depicting diverse weather conditions}
	\label{fig:dataset}
\end{figure}

Figure \ref{fig:dataset} shows the overview of the generated \acs{sotif}-related dataset in form of images. 
The dataset consists of 547 frames, primarily allocated for performance testing of a pre-trained \acs{dl} models, with 492 frames dedicated to the test set and 55 frames to the validation set.

\subsection{Application of 3D Object Detection Methods}
\label{subchap:application_3d_detection}
The computational system used was equipped with an Nvidia Quadro RTX 3000 GPU and an Intel Core i7-10850H CPU, with 64 GB of RAM, operating on both Microsoft Windows and Ubuntu, due to different software compatibility requirements.

MMDetection3D and OpenPCDet were chosen for their collection of pre-trained \acs{dl} models, a broad range of \acs{sota} 3D object detection methods, standardized benchmarking framework, and \acs{dl} model files tailored for simulation-based generated custom datasets.

The focus is on evaluating the adaptability and performance of \acs{sota} 3D object detection by using pre-trained models, trained on the KITTI dataset and testing them against a simulation-based generated dataset. Pre-trained models offer the advantage of saving substantial time and computational resources by utilizing networks that have been trained on extensive and diverse datasets.

Table \ref{tab:methods} provides a detailed comparison of the applied \acs{sota} 3D object detection methods. It highlights the methods, their data representation, feature extraction techniques, detection architectures, and the toolkits that support them.
\begin{table}[h!]
	\renewcommand{\arraystretch}{2} 

	\centering
	\caption{\acs{sota} 3D Object Detection Methods}
	\label{tab:methods}
	\begin{adjustbox}{width=7cm} 
	\begin{tabularx}{\textwidth}{|X|X|X|X|X|} 
			\hline
			
			\textbf{Methods} & 
			\textbf{\begin{tabular}[c]{@{}c@{}}Data\\Representation\end{tabular}} & 
			\textbf{\begin{tabular}[c]{@{}c@{}}Feature\\Extraction\end{tabular}} & 
			\textbf{\begin{tabular}[c]{@{}c@{}}Detection\\Architecture\end{tabular}} & 
			\textbf{\begin{tabular}[c]{@{}c@{}}Supported\\Toolkit\end{tabular}} \\\hline 

	\ac{part-a2} \newline \cite{shi2019part}   & Point-based & Point-by-point  & dual-stage & MMDetection3D and OpenPCDet \\ \hline 
			
	\ac{pv-rcnn} \newline \cite{shi2021pvrcnn}  & Voxel-based  & Segment-wise  &  dual-stage & MMDetection3D \\ \hline  
	
\ac{PointPillars} \newline \cite{lang2019pointpillars}  & Pillar-based  & \ac{cnn} & single-stage &  MMDetection3D and OpenPCDet \\ \hline 

	\ac{mvx-net} \newline \cite{sindagi2019mvxnet} &  Voxel-based & \ac{cnn} &  dual-stage &  MMDetection3D \\ \hline 
			
Dynamic Voxelization \newline \cite{zhou2019endtoend} & Voxel-based & \ac{cnn} & single-stage & MMDetection3D \\ \hline 
			
\ac{second}  \newline \cite{Yan2018SECONDSE} & Voxel-based& \ac{cnn} & single-stage &  MMDetection3D and OpenPCDet \\ \hline 

\ac{pointrcnn} \newline \cite{shi2019pointrcnn} & Point-based & Point-by-point & dual-stage & OpenPCDet \\ \hline 
		\end{tabularx}
	\end{adjustbox}
\end{table}

However, the application of 3D object detection methods to simulation-based datasets can introduce challenges. Factors including the domain gap between simulated and real-world scenarios, discrepancies in sensor characteristics, and variations in data distribution lead to mismatches in data characteristics and potentially impede the accuracy of the \acs{dl} models.

\begin{figure}[h!]
	\centering \includegraphics[width=\linewidth, keepaspectratio]{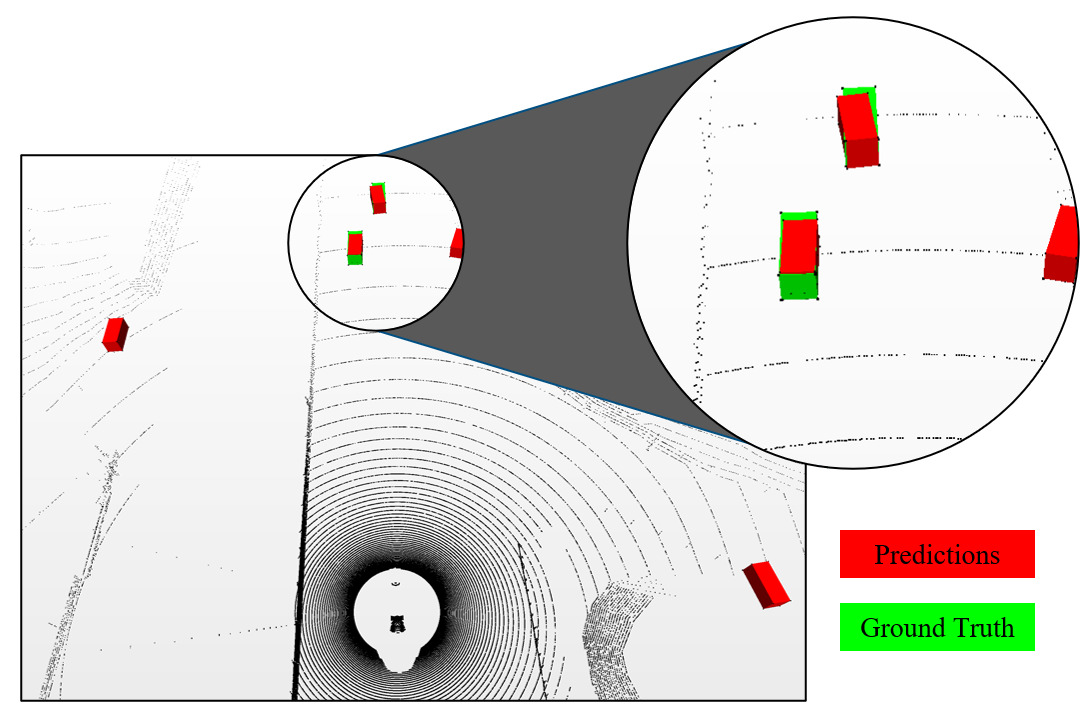}
	\caption{Visualisation of point cloud data and 3D object detection results in MeshLab}
	\label{fig:visualize_results}
\end{figure}

Figure \ref{fig:visualize_results} displays a point cloud visualization, demonstrating the application of an object detection method within the MeshLab tool.

The point cloud, illustrated as a grayscale backdrop of lines and dots, represents the environment captured by simulation. Within this environment, red boxes highlight the objects as predicted by the detection method, while green boxes illustrate their verified actual positions, referred to as the ground truth. The performance of the object detection method is determined by the degree of overlap between the red and green boxes.

\section{\uppercase{Performance Evaluation Results}}
\label{chap:results_discussion}
This chapter compares the performance of 3D object detection methods applied to generated dataset from a SOTIF-related Use Case, using MMDetection3D and OpenPCDet toolkits.

\subsection*{MMDetection3D}
The Table \ref{tab:mmdet_results} shows the performance comparison of 3D object detection methods  across \enquote{easy}, \enquote{moderate}, and \enquote{hard} difficulty levels, measured by AP$_{11}$ and AP$_{40}$ at an \acs{iou} of $0.70$. \cite{Geiger.2012} 

\begin{table}[!htbp]
	\renewcommand{\arraystretch}{2.5} 
	\centering
	\caption{\acs{ap} performance comparison of 3D object detection methods using MMDetection3D}
	\label{tab:mmdet_results}

	\begin{adjustbox}{max width=7cm}
			\begin{tabular}{|c|c|c|c|c|c|c|}
				\hline
				\multirow{2}{*}{\textbf{Method}} & \multicolumn{3}{c|}{\textbf{AP$_{11}$, (\acs{iou}=0.70)}} & \multicolumn{3}{c|}{\textbf{AP$_{40}$, (\acs{iou}=0.70)}} \\
				\cline{2-7}
				& \textbf{Easy} & \textbf{Moderate} & \textbf{Hard}  & \textbf{Easy} & \textbf{Moderate} & \textbf{Hard}  \\
				\hline
				\acs{part-a2} & 85.5935 & 75.9445 & 75.4239  & 86.5662 & 77.8658 & 75.6141  \\
				\hline
				\acs{pv-rcnn} & 89.7738 & 88.0915 & 87.3886  & 89.7959 & 82.2261 & 79.5364  \\
				\hline
				\acs{PointPillars} & 86.2029 & 76.9022 & 74.0742  & 94.8259 & 90.9872 & 87.7803  \\
				\hline
				\acs{mvx-net} & 81.9914 & 70.9114 & 71.7628  & 82.8947 & 70.6838 & 70.3484  \\
				\hline
				Dynamic Voxelization & 89.4193 & 87.8190 & 85.8189  & 93.2262 & 80.2123 & 70.8684  \\
				\hline
				\acs{second} & 87.0021 & 76.9475 & 74.8431  & 88.5588 & 81.4183 & 75.3544  \\
				\hline
			\end{tabular}
		\end{adjustbox}
	\end{table} 

The comparison shows that \acs{pv-rcnn} and \acs{PointPillars} exhibit higher \acs{ap} values, in \enquote{easy} conditions.

\subsection*{OpenPCDet}
The Table \ref{tab:openpcdet_results}, presents performance comparison of 3D object detection methods using the Recall metric at \acs{iou} thresholds of $0.30$ and $0.50$ for \acs{roi} and \acs{rcnn} stages. 

\begin{table}[!htbp]
	\caption{Recall performance comparison of 3D object detection methods using OpenPCDet }
		\label{tab:openpcdet_results}
		\renewcommand{\arraystretch}{2}
		\footnotesize
		\centering
		\begin{adjustbox}{max width=7cm}
			\begin{tabular}{|c|c|c|c|c|}
				\hline
				\multirow{2}{*}{\textbf{Method}} & \multicolumn{2}{c|}{\textbf{Recall, (\acs{iou}=0.30)}} & \multicolumn{2}{c|}{\textbf{Recall, (\acs{iou}=0.50)}} \\
				\cline{2-5}
				& \textbf{\acs{roi}} & \textbf{\acs{rcnn}} & \textbf{\acs{roi}} & \textbf{\acs{rcnn}} \\
				\hline
				\acs{part-a2} & 0.516 & 0.515 & 0.354 & 0.346 \\
				\hline
				\acs{pointrcnn} & 0.450 & 0.460 & 0.232 & 0.288 \\
				\hline
				\acs{second} & 0.515 & 0.5158 & 0.369 & 0.369 \\
				\hline
			\end{tabular}
		\end{adjustbox}
	\end{table}
The results maintain a consistent recall rate above 50\% at the lower \acs{iou} threshold of 0.30, with a noticeable decline at the higher threshold of 0.50.

The implications of these results are twofold. First, they confirm the feasibility of using pre-trained models on simulation-based generated custom dataset. Second, the difference in performance across difficulty levels and \acs{iou} thresholds underscores the need for further optimization and potential customization of \acs{dl} models to bridge the gap between simulated and real-world data accuracy. 
	
\section{\uppercase{Conclusion and Future Work}}
\label{chap:conclusion}
This paper investigates the application of 3D object detection methods in a Use Case related to Safety Of The Intended Functionality (SOTIF). The methodology and implementation, detailed in chapters \ref{chap:methodology} and \ref{chap:implementation}, address RQ1 and RQ2 (described in sub-chapter \ref{RQ1}).

The approach to preparing a dataset from a SOTIF-related Use Case simulation involved using the CARLA simulation environment to model a SOTIF-related Use Case, capturing 21 diverse weather conditions. The resultant dataset, structured in KITTI format, ensures compatibility and adaptability with State-of-the-Art (SOTA) 3D object detection methods designed for LiDAR point cloud datasets using Deep Learning (DL) models.

MMDetection3D and OpenPCDet toolkits were employed to assess the compatibility of \acs{sota} 3D object detection methods, including \acs{part-a2}, \acs{pv-rcnn}, \acs{PointPillars}, \acs{mvx-net}, Dynamic Voxelization \acs{second}, and \acs{pointrcnn}. Performance was benchmarked using pre-trained DL models, initially trained on the KITTI dataset, against the dataset generated from the SOTIF-related Use Case simulation.

Chapter \ref{chap:results_discussion} presents the performance evaluation of these methods, using Average Precision (AP) and recall metrics. The results demonstrate the effectiveness of these methods, offering detailed insights into their performance across different \acs{iou} thresholds, thereby addressing RQ3. Based on the results, a key recommendation is to focus on optimizing and customizing \acs{dl} models to improve their performance on simulated datasets. Adding more complex scenarios to the dataset could help reduce the performance gap between simulated and real-world data.

For future work, evaluating the uncertainty in 3D object detection models is a promising direction. This approach involves quantifying the confidence of \acs{dl} models in their predictions. Incorporating uncertainty evaluation allows for the identification of areas where \acs{dl} models are less reliable, guiding efforts to improve data, models, and training processes. Focusing on uncertainty can enhance model interpretability and trustworthiness, making it particularly valuable in scenarios where making the correct decision is critical.

\bibliographystyle{apalike}
{\small
\bibliography{references}}

\section*{\uppercase{Appendix}}

\subsection*{Supplementary Data}
The dataset and the associated script is available at the following link: \url{https://dx.doi.org/10.21227/j43q-z578} 

\subsection*{Abbreviations}
\label{chap_abbreviations}
\begin{acronym}
	\acro{ads}[ADS]{Automated Driving System}
	\acrodefplural{ads}[ADSs]{Automated Driving Systems}
	\acro{ap}[AP]{Average Precision}
	\acro{cnn}[CNN]{Convolutional Neural Network}
	\acrodefplural{cnn}[CNNs]{Convolutional Neural Networks}
	\acro{dl}[DL]{Deep Learning}
	\acro{FN}[FN]{False Negative}
	\acro{FP}[FP]{False Positive}
	\acro{iou}[IoU]{Intersection over Union}
	\acro{lidar}[LiDAR]{Light Detection And Ranging}	
	\acro{mvx-net}[MVX-Net]{Multimodal VoxelNet for 3D Object Detection}
	\acro{part-a2}[Part-A\textsuperscript{2}]{Part-aware and Part-aggregation Network}
	\acro{PointPillars}[PointPillars]{Fast Encoders for Object Detection from Point Clouds}
	\acro{pointrcnn}[PointRCNN]{3D object proposal generation and detection from point}
	\acro{pv-rcnn}[PV-RCNN]{Point-Voxel Feature Set Abstraction for 3D Object Detection}
	\acro{rcnn}[R-CNN]{	Region-based Convolutional Neural Networks }
	\acro{roi}[RoI]{Region of Interest}
	\acro{rpn}[RPN]{Region Proposal Network}
	\acro{second}[SECOND]{Sparsely Embedded Convolutional Detection}
	\acro{sota}[SOTA]{State-of-the-Art}
	\acro{sotif}[SOTIF]{Safety Of The Intended Functionality}
	\acro{tcp}[TCP]{Transmission Control Protocol}
	\acro{TN}[TN]{True Negative}
	\acro{TP}[TP]{True Positive}
	\acro{v2v}[V2V]{Vehicle-to-Vehicle}
\end{acronym}

\end{document}